%% file: main.tex
\documentclass[conference]{IEEEtran}
\IEEEoverridecommandlockouts
\usepackage{cite}
\usepackage{amsmath,amssymb,amsfonts}
\usepackage{algorithm2e}
\usepackage{subcaption}

\usepackage{graphicx}
\usepackage{textcomp}
\usepackage{xcolor}
\usepackage{url}
\def\BibTeX{{\rm B\kern-.05em{\sc i\kern-.025em b}\kern-.08em
    T\kern-.1667em\lower.7ex\hbox{E}\kern-.125emX}}

\newcommand{\acronym}{ContextGPT}

\begin{document}

\title{ContextGPT: Infusing LLMs Knowledge into
Neuro-Symbolic Activity Recognition Models\\
}

\author{\IEEEauthorblockN{Luca Arrotta, Claudio Bettini, Gabriele Civitarese, Michele Fiori}
\IEEEauthorblockA{\textit{EveryWare Lab, Dept. of Computer Science} \\
\textit{University of Milan}\\
Milan, Italy \\
\{luca.arrotta, claudio.bettini, gabriele.civitarese, michele.fiori\}@unimi.it}

}

\maketitle

\begin{abstract}
Context-aware Human Activity Recognition (HAR) is a hot research area in mobile computing, and the most effective solutions in the literature are based on supervised deep learning models. However, the actual deployment of these systems is limited by the scarcity of labeled data that is required for training. Neuro-Symbolic AI (NeSy) provides an interesting research direction to mitigate this issue, by infusing common-sense knowledge about human activities and the contexts in which they can be performed into HAR deep learning classifiers. Existing NeSy methods for context-aware HAR rely on knowledge encoded in logic-based models (e.g., ontologies) whose design, implementation, and maintenance to capture new activities and contexts require significant human engineering efforts, technical knowledge, and domain expertise.  
Recent works show that pre-trained Large Language Models (LLMs) effectively encode common-sense knowledge about human activities. In this work, we propose \acronym: a novel prompt engineering approach to retrieve from LLMs common-sense knowledge about the relationship between human activities and the context in which they are performed. Unlike ontologies, \acronym\ requires limited human effort and expertise, while sharing similar privacy concerns if the reasoning is performed in the cloud. An extensive evaluation using two public datasets shows how a NeSy model obtained by infusing common-sense knowledge from \acronym\ is effective in data scarcity scenarios, leading to similar (and sometimes better) recognition rates than logic-based approaches with a fraction of the effort.
\end{abstract}

\begin{IEEEkeywords}
human activity recognition, context-awareness, large language models
\end{IEEEkeywords}

\section{Introduction}
\input{sections/introduction}

\section{Related Work}

\input{sections/related}


\section{Methodology}

\input{sections/method}

\section{Experimental Evaluation}

\input{sections/experiments}

\section{LLM vs. Ontology for Neuro-Symbolic HAR}
\input{sections/discussion}


\section{Conclusions and Future Work}

\input{sections/conclusion}

\section*{Acknowledgements}
The authors thank Mochen Laaroussi for his excellent implementation work.
This work was supported in part by projects FAIR (PE00000013) and SERICS (PE00000014) under the NRRP MUR program funded by the EU - NGEU. Views and opinions expressed are however those of the authors only and do not necessarily reflect those of the European Union or the Italian MUR. Neither the European Union nor the Italian MUR can be held responsible for them.

\bibliographystyle{unsrt}
\bibliography{references.bib}

\end{document}

%% file: sections/introduction.tex
%
The analysis of sensor data gathered from mobile and wearable devices for Human Activity Recognition (HAR) has been extensively researched~\cite{chen2021deep,gu2021survey}. This is attributed to its high potential for applications across various domains such as well-being, healthcare, and sports, fueled by the widespread adoption of these devices.
Although the majority of the studies in this area focused on inertial sensors only, context-aware HAR approaches also take into account the user's surrounding context as derived from the devices, user profiles or online services (e.g., semantic location, temporal information, etc.) to further improve the recognition rate and, at the same time, extend the set of recognizable activities~\cite{bettini2020caviar}.

In the last few years, a vast amount of solutions based on deep learning classifiers have been proposed~\cite{wang2019deep}. However, most of those approaches rely on supervised learning and require large labeled datasets to be trained. 
The need to acquire reliable labels from a large number of users and for an extended set of activities is currently a major obstacle to the deployment of effective sensor-based HAR systems.

Among the several solutions proposed in the general machine learning community to tackle the labeled data scarcity problem, Neuro-Symbolic (NeSy) approaches represent a promising research direction~\cite{sarker2021neuro}. NeSy methods aim at combining data-driven and knowledge-based approaches to reduce the amount of labeled data needed to effectively train the model and, at the same time, to improve its interpretability. In NeSy approaches, well-known domain constraints are infused into the model, thus avoiding learning them from data and simplifying the overall learning process.

Existing NeSy approaches proposed for context-aware HAR retrieve common-sense knowledge from logic-based models (e.g., ontologies) manually designed and implemented by human experts~\cite{arrotta2022knowledge}. Such knowledge models encode the relationship between an activity and the context in which it can be performed.
For instance, they 
encode that \textit{biking} is not a plausible activity if the user's semantic location is \textit{highway} or \textit{museum}. However, manually building a comprehensive knowledge model that captures all possible context situations is challenging, and, even relying on a domain expert, it does not guarantee that all the possible situations in which an activity can be performed are captured. Moreover, logic-based models are not scalable since including new activities, new context conditions, or new constraints requires further manual work and expertise.

This paper explores the idea of infusing common-sense knowledge into NeSy HAR models from Large Language Models (LLMs) instead of ontologies.
%
%
Indeed, LLMs implicitly encode knowledge spanning a wide range of domains and recent results suggest that they can be efficiently exploited to retrieve common-sense knowledge about human activities~\cite{takeda2023sensor,graule2023gg,zhou2023tent,kaneko2023toward,xia2023unsupervised,leng2023generating}. 

In this paper, we propose \acronym, a novel prompt engineering approach that, based on the high-level user's context, 
leverages a pre-trained LLM model to determine the most plausible activities.
Specifically, given a time window of sensor data, \acronym\ converts high-level context information into a natural language description. Thanks to a carefully designed system message, \acronym\ generates a prompt by asking the LLM to determine the activities that are consistent with the current context.
The prompts generated by \acronym\ also include a few examples (created by the prompt engineer) depicting how the task should be carried out (i.e., few-shot prompting). 



Our contributions are the following:
\begin{itemize}
    \item We propose \acronym: a novel prompt engineering approach to retrieve common-sense knowledge about the relationships between activities and high-level contexts from pre-trained Large Language Models.
    \item We illustrate how this knowledge can be infused into a Neuro-Symbolic Context-Aware model to mitigate labeled data scarcity.
    \item Our extensive experimental evaluation using two public datasets shows that infusing \acronym\ knowledge leads to recognition rates similar to methods based on logic-based models, while significantly reducing the cost in terms of expert human effort.
\end{itemize}

%% file: sections/related.tex
\label{sec:related}

\subsection{Data scarcity in HAR}



Due to the unrealistic amount of training data required to train supervised models, several research groups are proposing solutions to leverage small amounts of labeled samples. Proposed methods to mitigate labeled data scarcity are based on transfer learning~\cite{dhekane2024transfer}, self-supervised learning~\cite{haresamudram2022assessing}, and semi-supervised learning approaches~\cite{abdallah2018activity}. 
We believe that Neuro-Symbolic AI (NeSy) could be coupled with such approaches to further mitigate data scarcity when fine-tuning pre-trained HAR models with limited labeled data.


\subsection{Neuro-Symbolic HAR}
While several NeSy methods have been proposed in the computer vision and NLP domains~\cite{dash2022review}, only a few approaches have been proposed for HAR.


Existing NeSy methods for context-aware HAR retrieve common-sense knowledge from logic-based models (e.g., ontologies). To the best of our knowledge, three main strategies have been proposed so far to combine extracted knowledge with deep learning models: a) using knowledge to refine the deep model's output~\cite{bettini2020caviar}, b) including retrieved knowledge as additional features in the latent space~\cite{arrotta2020context}, and c) using a loss function that penalizes predictions violating domain constraints~\cite{arrotta2024semantic, xing2020neuroplex}. 
However, designing and implementing knowledge models require significant human effort, and those models may not capture all the possible situations in which activities can be performed.
While there are information retrieval approaches to semi-automatically obtaining common-sense knowledge from public external sources (e.g., images~\cite{riboni2019sensor}, web~\cite{perkowitz2004mining}, text~\cite{yordanova2016textual}) such methods still faces challenges in creating comprehensive knowledge models.


\subsection{Using LLMs for Human Activity Recognition}
The adoption of Large Language Models (LLMs) in sensor-based HAR is a recently emerging trend, and we expect a surge of contributions in this area in the near future. For instance, the work in~\cite{takeda2023sensor} takes advantage of the GPT-2 model to predict sequences of sensor events in smart-home settings. Specifically, since the GPT-2 model is pre-trained to predict the next word in a sentence, it has been fine-tuned to predict the most likely sequence of sensor events that are treated similarly to textual data. A similar approach was also proposed in~\cite{graule2023gg}, leveraging the LLaMA2 language model. The work in~\cite{zhou2023tent} uses the CLIP pre-trained model to align sensor data embeddings and text embeddings by means of contrastive learning, with the goal of improving the overall recognition rate.

A few research groups recently proposed solutions based on the well-known ChatGPT tool. In~\cite{kaneko2023toward}, ChatGPT is prompted with the description of a specific sensor-based HAR task and the current choice for sensor positioning, with the objective of obtaining feedback about where to install new sensors to further improve the recognition rate. The work in~\cite{xia2023unsupervised} proposed an unsupervised approach for smart-home settings. The approach consists of two steps: the former asks ChatGPT to provide a description of the target activities, and the latter asks ChatGPT to infer the performed activity given a temporal sequence of events describing interaction with household items. 
Finally, based on the recent advances in generating synthetic inertial sensor data from textual descriptions~\cite{guo2022generating}, the work in~\cite{leng2023generating} leverages ChatGPT to generate textual descriptions of activities, that are then used to generate virtual inertial sensors data.

To the best of our knowledge, we are the first leveraging LLMs and NeSy models to mitigate data scarcity in context-aware HAR.

%% file: sections/method.tex
\label{sec:methodology}

\subsection{A Neuro-Symbolic Framework for Context-Aware HAR}
\begin{figure}[h!]
\centering
\includegraphics[width=0.7\columnwidth]{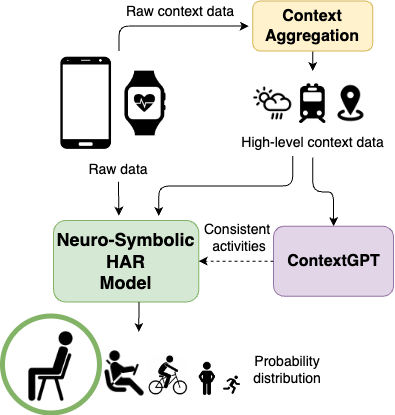}
\caption{A Neuro-Symbolic AI framework for context-aware HAR gathering knowledge from \acronym}
\label{fig:arch}
\end{figure}

We consider a mobile and wearable computing scenario in which users perform activities while carrying their devices (e.g., smartphone, smartwatch, tracker).
Figure~\ref{fig:arch} provides an overview of the Neuro-Symbolic (NeSy) framework taking as input the continuous stream of sensor data generated by the user devices, and providing as output the most likely performed activity. 

Data are partitioned in fixed-size temporal windows, and for each window $w$ we derive two subsets $w^R$ and $w^C$: the former includes raw data that we consider appropriate for being directly processed by a data-driven model (e.g., inertial sensors data), while the latter involves raw sensor data that we consider useful for deriving high-level contexts through reasoning and/or abstraction. 

A high-level context provides information, at a certain point in time, about the user, the environment surrounding them, and their interactions with objects or other subjects. 
Let $C = \langle C_1, \dots, C_n \rangle$ be a set of possible contexts that are meaningful for the application domain (e.g., $C_1$ = \textit{it is raining}, $C_2$ = \textit{location is a park}, $C_3$ = \textit{current user's speed is high}).

Note that $w^R$ and $w^C$ may have a non-empty intersection and $w^R \bigcup w^C = w$. For instance, it may be appropriate to exclude raw GPS data from $w^R$ since it may be difficult to find robust correlations capable of generalizing between different users (especially in data scarcity settings). On the other hand, raw GPS data can be included in $w^C$ to obtain high-level contexts that are more easily correlated with activities (e.g., semantic location: "public park").

The \textit{Context Aggregation} module is in charge of deriving all the most likely high-level contexts $C^w \subset C$ that occur in a window $w$ based on $w^C$.  \textit{Context Aggregation} can be implemented using simple rules, available services, and/or context-aware middlewares. For example, raw GPS coordinates can be used to derive the semantic location by querying a dedicated web service (e.g., by using Google Places APIs).

\textit{Context-aware HAR} could be tackled with machine learning models taking $\langle w^R, C^w \rangle$ only as input. However, a more effective approach is to complement data-driven approaches with common-sense knowledge about the relationships between activities and contexts~\cite{riboni2011cosar}. For example, people typically run outdoors in parks, on trails, or along sidewalks (preferably in dry weather) and indoors on a treadmill. Moreover, running requires the user to move at a positive speed.
The relationships between activities and the typical conditions in which they can be performed can be used in the HAR process to reduce the amount of labeled data required to learn them.

The \acronym\ module is in charge of reasoning on the contexts in $C^w$ to derive the most likely activities that are consistent according to $C^w$. \acronym\ relies on a Large Language Model (LLM) and it is described in detail in the rest of this section. The information about context-consistent activities is infused into a NeSy HAR model, which also receives as input raw data and high-level context data (i.e. $\langle w^R, C^w \rangle$). The output of the model is a probability distribution $P = \langle p_1, \dots, p_k \rangle$, where $p_i$ is the probability that the user performed the activity $a_i$.

\baselineskip=12bp
\subsection{ContextGPT architecture}

\begin{figure}[h!]
\centering
\includegraphics[width=0.7\columnwidth]{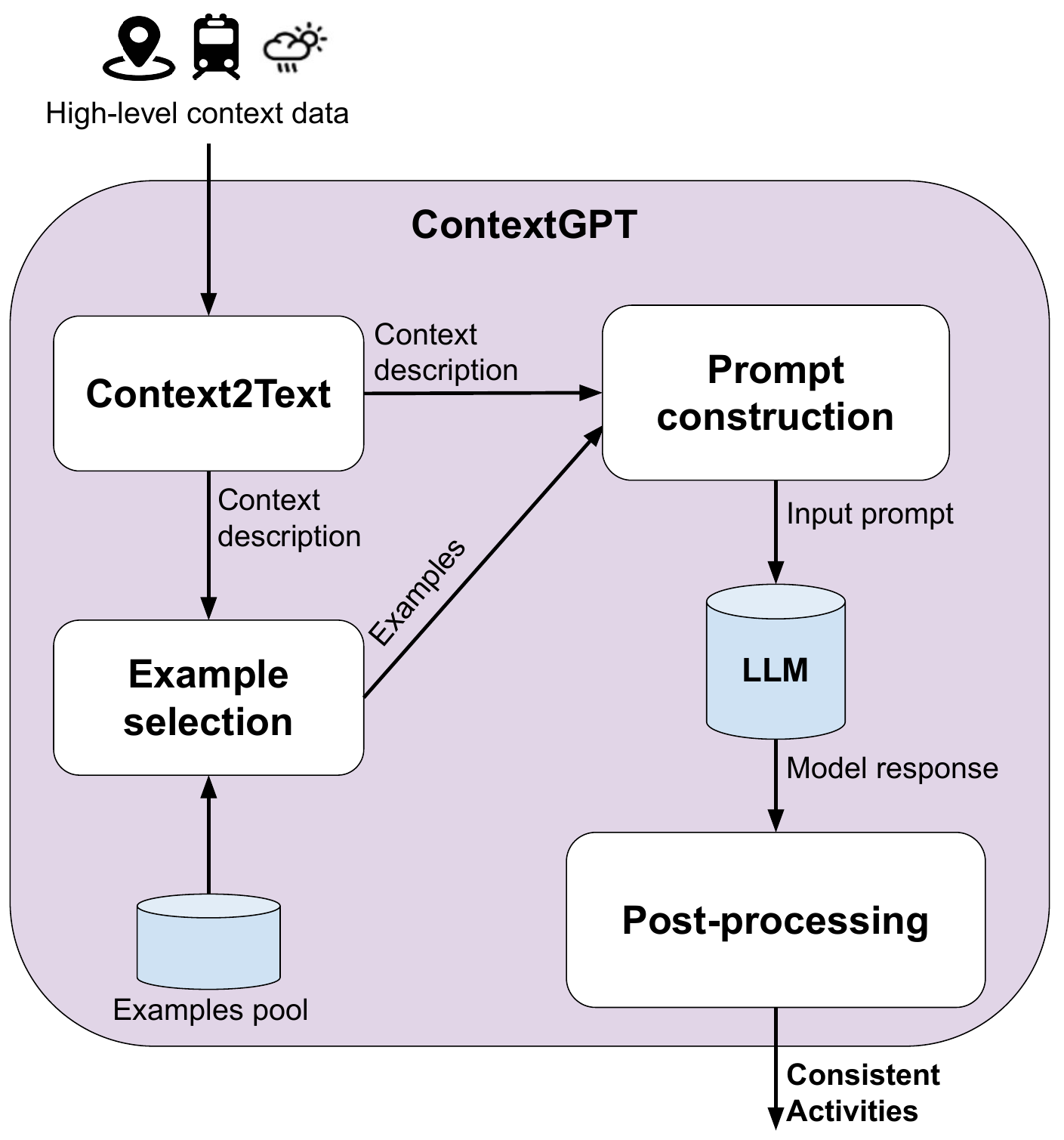}
\caption{Overall architecture of \acronym}
\label{fig:contextgpt}
\end{figure}

Figure~\ref{fig:contextgpt} illustrates the overall architecture of \acronym.

\acronym\ receives as input a temporal window of high-level context data $C^w$. First, $C^w$ is provided to the \textsc{Context2Text} module to obtain a natural language description of the user's context. Since LLMs also benefit from a few examples depicting how the required task should be carried out (i.e., the so-called \textit{few-shot prompting}~\cite{brown2020language}), the \textsc{Example selection} module 
considers a pool of examples and includes in the prompt those having their context similar to $C^w$. Each example depicts a context situation and the activities that are consistent with that situation. Finally, the \textsc{Prompt Construction} module generates the complete prompt that is composed of:
\begin{itemize}
    \item \textbf{System Message}: general instructions to the LLM about the task that it has to carry out.
    \item \textbf{Examples}: the most similar examples to the input context, selected from a pool.
    \item \textbf{Context Description}: the description in natural language of $C^w$.
\end{itemize}
The prompt is provided as input to a pre-trained LLM, and the output is post-processed to obtain the list of activities that are consistent with the current context.

\subsection{Prompt Engineering}

While existing Neuro-Symbolic methods for Context-Aware HAR demand knowledge engineers to manually build logic-based models, \acronym\ requires a Prompt Engineer in charge of designing: i) the system message, ii) the translation of context data into natural language descriptions, and iii) the examples in the pool. Due to the sensitivity of LLMs to the specific wording in the prompt, the Prompt Engineer currently has a non-trivial role for the success (or failure) in obtaining the desired goal~\cite{meyer2023chatgpt}. However, since these tasks are based on natural language, there is no need for designing comprehensive and complex relationships between activities and contexts using logic formalisms; hence, the required expertise and human effort is significantly reduced. In the following, we describe each component of the prompts of \acronym\ in detail.

\subsubsection{System Message}
\label{subsec:systemprompt}

\begin{figure}[h!]
\centering
\includegraphics[width=\columnwidth]{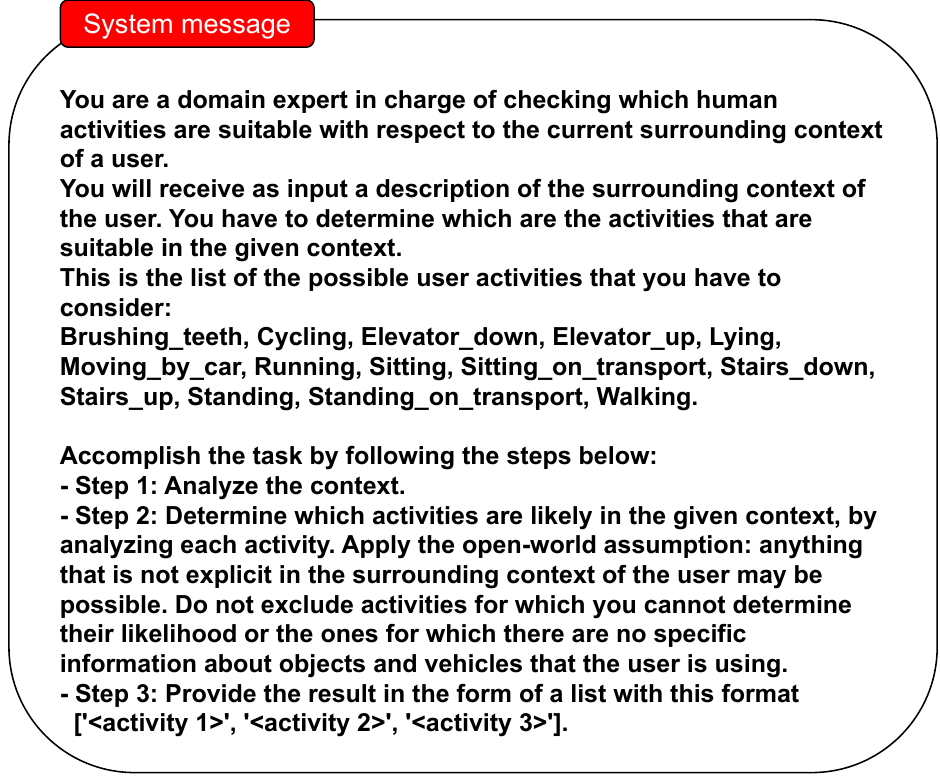}
\caption{The system message of \acronym. The possible activities, in this case, are the ones of the DOMINO~\cite{arrotta2023domino} dataset.}
\label{fig:systemprompt}
\end{figure}

The system message defines the general description of the task the LLM should complete. In our case, the task is determining the activities that are consistent with a given context. Hence, we divided the system message into two parts. The former instructs the LLM about the overall task and the list of possible activities. The latter provides a detailed list of steps the LLM should undertake to complete the task (i.e., Chain-Of-Thought approach~\cite{wei2022chain}).
The first step directs the LLM to focus on the input context. The second step requires following an open-world assumption since, in our preliminary experiments, we noticed instances where the model mistakenly excluded activities not explicitly supported by the context description. For instance, consider the following context description: \textit{``In the last 4 seconds the user Bob was in an outdoor environment, where he was moving/traveling at speed between 1 and 4 km/h, experiencing rainy weather, not following/close to a public transportation route, and not experiencing elevation changes.''}. Without the second step, in our preliminary experiments, the LLM often excluded the activity \textit{Moving by Car} with the following motivation: \textit{``Not consistent as there is no mention of being in a car or any other vehicle besides walking speed movement''}. While it is true that being in a car is not explicitly provided in the context description, the \textit{Moving by car} activity should still be considered as possible. Indeed, it is impractical to monitor all the possible contextual conditions through sensors in mobile/wearable devices. Finally, the last step forces the LLM to provide context-consistent activities in a specific format to simplify the automatic extraction of this information during post-processing. Figure~\ref{fig:systemprompt} shows the system message of \acronym. 

\subsubsection{Context2Text} 

In order to be ingested by the LLM, the \textsc{Context2Text} module transforms the input context data $C^w$ into a natural language sentence describing it. Each description starts with `\textit{`In the last $z$ seconds, the user $u$''} to contextualize that the described context refers to a specific temporal window of $z$ seconds and that it is associated with a specific user $u$.
Then, the sentence continues by describing in natural language each context variable. Designing the specific mapping between context data and natural language sentences is currently not trivial, since the prompt engineer has to understand (through trial and error) how the LLM interprets specific words. 
For instance, the context
\textit{``the user is on a public transportation route''} means that the path that the user is following (e.g., as captured by a GPS trace) is the one of a public transportation route. However, these words sometimes led the model to mistakenly interpret it as “the
user is on a public transportation vehicle”, thus excluding activities like \textit{Moving by car} and
\textit{Cycling}. We observed that translating the same context information as \textit{``the user is following/close to a
public transportation route''} significantly reduced the instances where this issue occurs.

\subsubsection{Example pool}

LLMs benefit from including in the prompt a few examples showing the output that should be associated with specific inputs. In our case, an example represents a context situation and the corresponding consistent activities. We assume that the Prompt Engineer is 
sufficiently knowledgeable in the activity domain and, given a context description, they are capable of determining which activities are consistent. For each activity $a$, the Prompt Engineer populates the pool of examples $P$ with the following strategy:

\begin{itemize}
    \item Define a set of examples $E_a$ (i.e. a set of tuples $\langle context$, $consistentActivities \rangle$) referring to context situations that are uncommon but possible for the activity $a$ (the context may be common or uncommon for the other consistent activities)
    \item For each example $e \in E_a$:
    \begin{itemize}
        \item Consider $e$ as an input context for the LLM (using the system message without examples) and analyze the response about the set of consistent activities.
        \item If the response of the LLM is different from the consistent activities in the example, the Prompt Engineer decides whether to include $e$ in $P$ to fill the knowledge gap.
    \end{itemize}
\end{itemize}

Considering one activity at a time significantly simplifies the Prompt Engineer's task of generating examples. The number of examples created by the prompt engineer for each activity is not fixed: it depends on the activity, the available contexts, and their experience in the activity domain. An example is added to the pool only if the Prompt Engineer feels that the LLM is not providing a satisfactory outcome, revealing a gap in the LLM knowledge. We consider ``uncommon'' context situations in which an activity is rarely performed but it is still possible; these are the cases most likely not covered by the LLM. For instance, consider the activity \textit{running}. While this activity is usually performed outdoors, it may be uncommon to think that it may also performed at the gym (e.g., on a treadmill). In our preliminary experiments, we observed that such uncommon context for \textit{running} was often not captured by the LLM, and including it as an example improved \acronym\ outputs significantly.

\subsubsection{Example Selection}
\label{subsec:examples}

Since the number of examples in the pool is not fixed, including all of them in the prompt may result in increased costs, especially when using external services whose price depends on the prompt's length. Moreover, there are often limits to the LLM's input size.

\acronym\ employs a method to include in the prompt only those examples that are similar to the input context $C^w$.
Specifically, we use a pre-trained LLM to extract text embeddings from $C^w$ and all the examples in the pool using their description in natural language obtained with \textsc{Context2Text}. Then, we use cosine similarity to compute the similarity of each example with $C^w$, and we include in the prompt only the examples with a similarity higher than a threshold $k$. This threshold is determined empirically, and we will show its impact on the recognition rate in Section~\ref{sec:experiments}.

\subsection{Post-processing}

\begin{figure}[h!]
\centering
\includegraphics[width=\columnwidth]{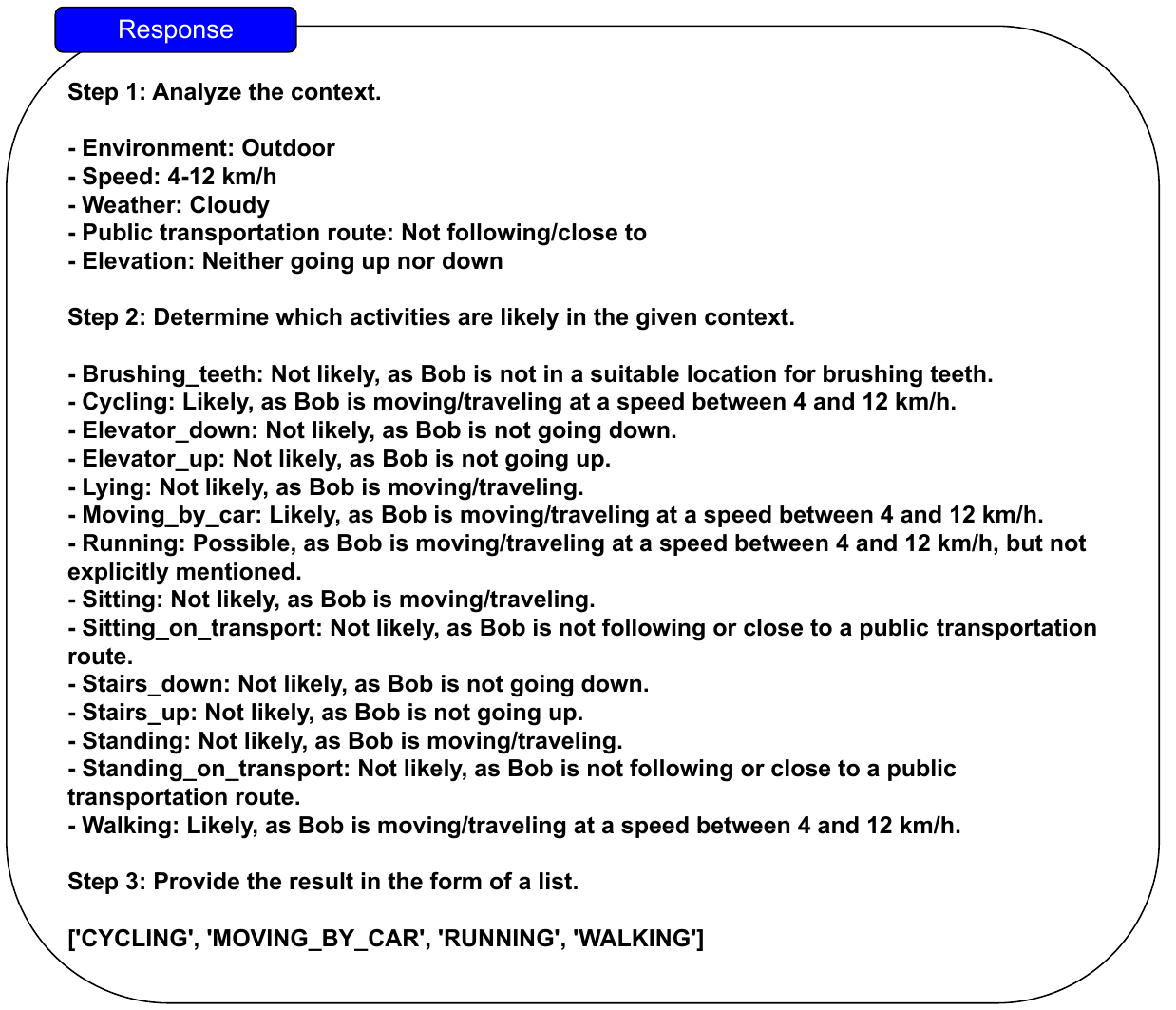}
\caption{An example of LLM output}
\label{fig:output}
\end{figure}

Figure~\ref{fig:output} shows an example of LLMs output in \acronym.

As required by the system message (Section~\ref{subsec:systemprompt}), besides explaining the reasoning process, the output includes the list of consistent activities in square brackets.
Hence, it is possible to easily extract this list $L$ using regular expressions and transform it into a binary vector $b = [b_1,b_2,\dots,b_n]$ where $b_i$ is $1$ if the activity $a_i \in L$ (i.e., the activity $a_i$ is consistent with $C^w$ according to the LLM), $0$ otherwise. 
In the following, we will refer to $b$ as the \textit{consistency vector}. 
Note that in the actual implementation of \acronym\ there is a cache mechanism that avoids recomputing the consistency vector when $C^w$ has already been processed.
Consistency vectors are used to infuse knowledge inside the NeSy model.

\subsection{Infusing knowledge into Neuro-Symbolic model}
\label{subsec:neuro}

While \acronym\ is agnostic with respect to the Neuro-Symbolic approach for Context-Aware HAR, in this work we use a specific knowledge infusion approach proposed in the general AI community~\cite{sheth2019shades,kursuncu2019knowledge}. This method relies on a hidden layer in the Deep Neural Network (DNN) in charge of infusing knowledge in the latent space. An adaptation for Context-Aware HAR named NIMBUS has been recently proposed in the literature, exhibiting promising recognition rates~\cite{arrotta2022knowledge}. In NIMBUS, knowledge is infused from a manually defined ontology of activity and contexts.
In this work, we adapted NIMBUS to fetch knowledge from \acronym.

\begin{figure}[h!]
\centering
\includegraphics[width=0.7\columnwidth]{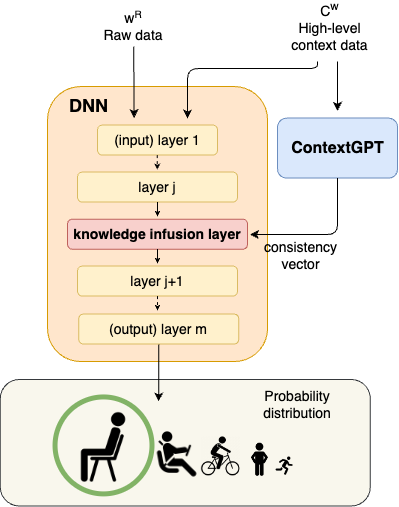}
\caption{Infusing \acronym\ into the Symbolic Features approach}
\label{fig:symbolicfeatures}
\end{figure}

The overall mechanism of NIMBUS is depicted in Figure~\ref{fig:symbolicfeatures}. The \textit{consistency vector} obtained from \acronym~ is infused in the hidden layers of the Deep Neural Network (DNN) through a dedicated layer named \textit{knowledge infusion layer}. This hidden layer makes it possible to learn correlations between the latent representation of raw sensor data, high-level context data, and context-consistent activities. 

%% file: sections/experiments.tex
\label{sec:experiments}

\subsection{Datasets}


\subsubsection{DOMINO}
The DOMINO~\cite{arrotta2023domino} dataset includes data from $25$ subjects wearing a smartwatch on the wrist of their dominant hand and a smartphone in their pants front pocket. Both devices gathered raw sensor data from inertial sensors and a wide variety of high-level context data. This dataset includes nearly $9$ hours of labeled data (approximately $350$ activity instances) covering $14$ distinct activities: \textit{walking}, \textit{running}, \textit{standing}, \textit{lying}, \textit{sitting}, \textit{stairs up}, \textit{stairs down}, \textit{elevator up}, \textit{elevator down}, \textit{cycling}, \textit{moving by car}, \textit{sitting on transport}, \textit{standing on transport}, and \textit{brushing teeth}.
The DOMINO dataset was collected in a scripted setting: the volunteers were instructed to perform sequences of indoor/outdoor activities, even though without specific guidance on their execution. 

DOMINO included several pre-processed high-level context information, like the semantic loation, weather conditoin, whether the user is following a public transportation route, user's height and speed variations, etcetera.
Overall, the DOMINO dataset includes data collected in $412$ unique context conditions.


\subsubsection{ExtraSensory} 

ExtraSensory~\cite{vaizman2018extrasensory} was collected in an unscripted in-the-wild setting from $60$ subjects.
Similarly to DOMINO, users wore a smartwatch on the wrist of their dominant hand and a smartphone in their pants front pocket.
ExtraSensory includes approximately $300$,$000$ minutes of labeled data, including $51$ distinct labels self-reported by the users. These labels encode both high-level context information (e.g., at home, with friends, phone in bag, phone is charging) and performed activities (e.g., sitting, bicycling).
Since it was collected in the wild, ExtraSensory is widely considered a challenging benchmark. Existing works on this dataset report low recognition rates although considering small subsets of activities~\cite{arrotta2024semantic,cruciani2020feature}.
In this paper, we pre-process the dataset consistently with previous works in context-aware HAR~\cite{arrotta2024semantic}. We consider the following $7$ activities: \textit{bicycling}, \textit{lying down}, \textit{moving by car,} \textit{on transport}, \textit{sitting}, \textit{standing}, and \textit{walking}. 
All the context information in the dataset that could be easily collected by mobile devices is provided as input to the Neuro-Symbolic model (e.g., audio level, screen brightness, ringer mode, etc.).  However, based on preliminary experiments, we provide to the LLM only the context information where common-sense knowledge can be practically used to reason about the target physical activities: the user's semantic location, their speed, movements diameters, and whether they are following a public transportation route.
Overall, the ExtraSensory data presents $144$ unique context conditions for the LLM. This number is significantly lower compared to DOMINO, due to a reduced number of context variables and target activities.




\subsection{Experimental setup}

We implemented a working prototype of \acronym\ as well as the Neuro-Symbolic model explained in Section~\ref{subsec:neuro} using the Python language (version 3.12). We run our experiments on a machine of our department running Ubuntu 20.04.4 LTS and equipped with an AMD EPYC Processor x86-64, an NVIDIA A100 GPU (80 GB), with $43.5$ GBs RAM allocated.
%
%
%

The pre-trained LLM we used in our experiments is \texttt{gpt-3.5-turbo} by OpenAI, queried through its API through the Python OpenAI package (version 0.28.1). We set the temperature of the model to $0$ to ensure a mostly deterministic output since our task does not require leveraging the model's creativity. The pre-trained LLM model we adopted to compute example embeddings and thus computing similarity (see Section~\ref{subsec:examples}) is \texttt{all-MiniLM-L6-v2}\footnote{\url{https://huggingface.co/sentence-transformers/all-MiniLM-L6-v2}}. 
Considering the NeSy model, we adopted the NIMBUS implementation proposed in~\cite{arrotta2024semantic}\footnote{Note that in~\cite{arrotta2024semantic} NIMBUS is named \textit{Symbolic Features}.} (with the same hyper-parameters).


%



Finally, a member of our research group (without experience with the specific datasets used in this work) created the examples for few-shot-prompting following the strategy described in Section~\ref{subsec:examples}. Since each dataset has different sets of activities and context variables, examples were created separately for DOMINO and ExtraSensory. The output of this process is $21$ examples for DOMINO and $12$ for ExtraSensory.


\subsection{Evaluation methodology}

\subsubsection{Baselines}

In this paper, we compare the infusion of \acronym\ knowledge into a Neuro-Symbolic HAR model with two alternatives:
\begin{itemize}
    \item \textbf{No knowledge}: a purely data-driven baseline without knowledge infusion, where high-level context data is only used as input of the deep learning model. 
    \item \textbf{Ontology}: the NIMBUS approach originally proposed in~\cite{arrotta2022knowledge}, where knowledge is infused from an ontology encoding relationships between activities and contexts. We use the ontology adopted in that work for the DOMINO dataset and its adaptation to the ExtraSensory dataset recently proposed in~\cite{arrotta2024semantic}.
\end{itemize}

\begin{figure*}
\centering
    \begin{subfigure}[b]{0.70\columnwidth}
        \includegraphics[width=\columnwidth]{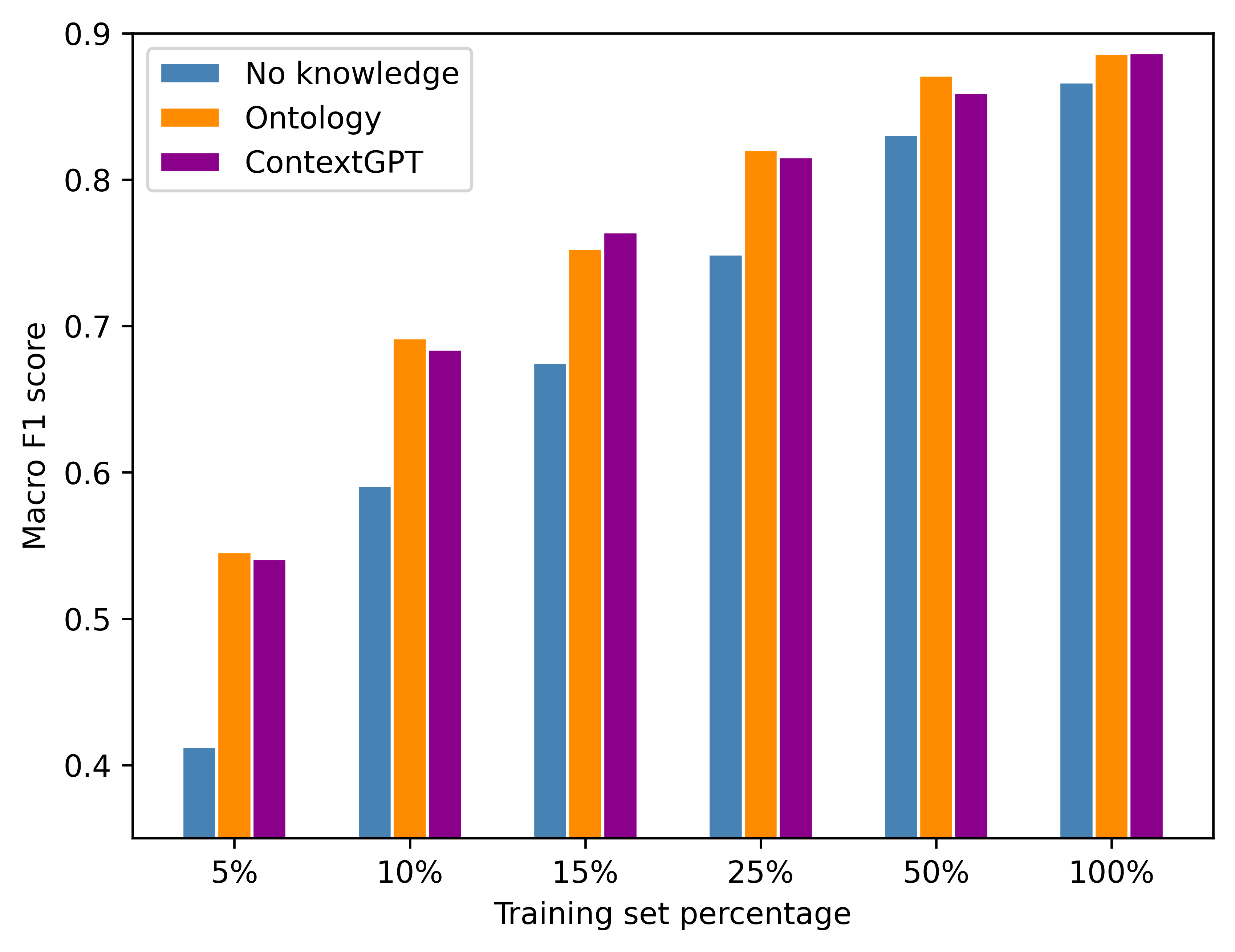}
        \caption{DOMINO}
        \label{fig:domino-main}
    \end{subfigure}%
    \begin{subfigure}[b]{0.70\columnwidth}
        \includegraphics[width=\columnwidth]{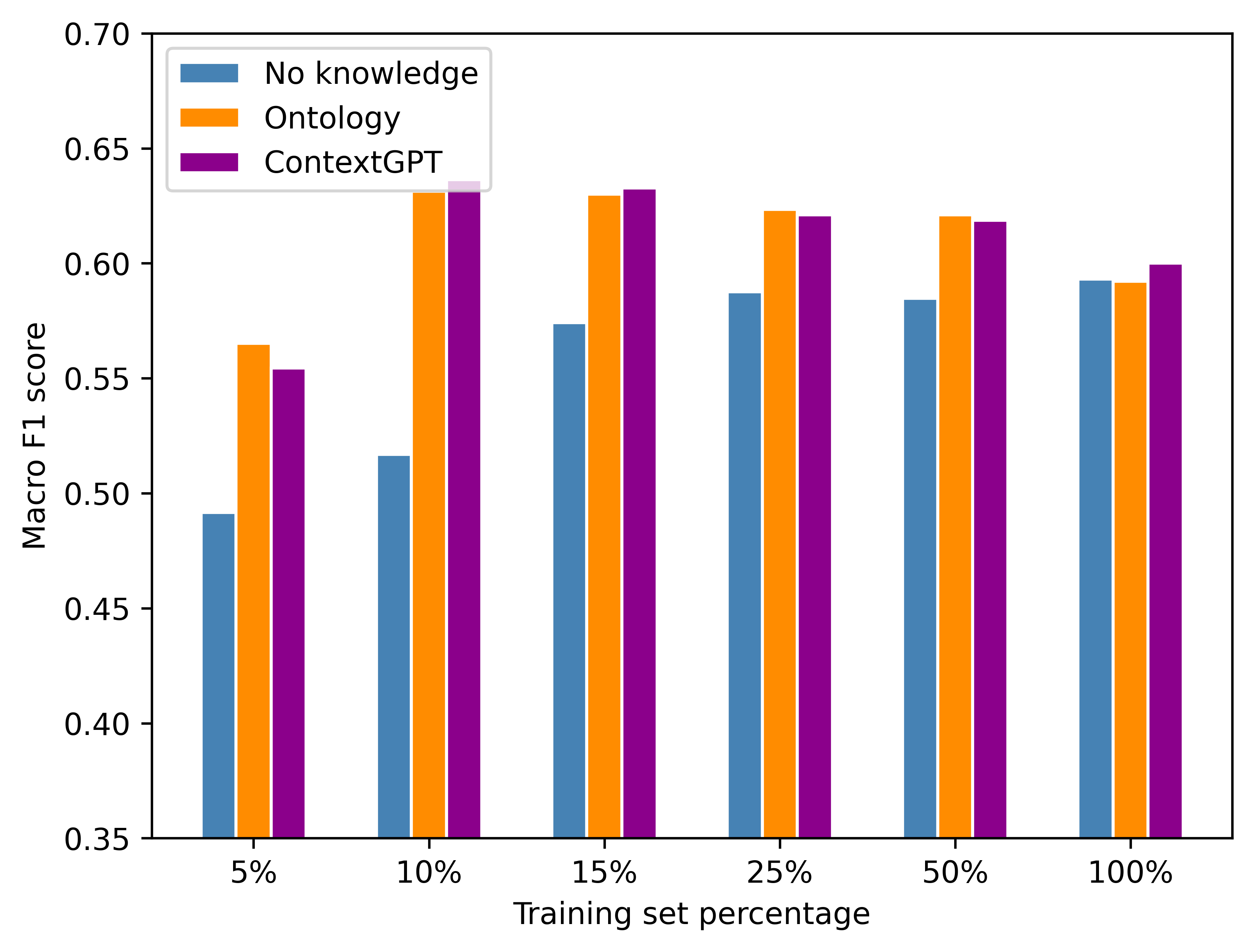}
        \caption{ExtraSensory}
        \label{fig:extrasensory-main}
    \end{subfigure}
    \caption{Knowledge infusion with \acronym\ compared to the baselines. This plot shows the F1 score in different data scarcity scenarios.}
    \label{fig:combined}
\end{figure*}

\subsubsection{Evaluation Strategy}

We evaluated the recognition rate of the HAR models by using a leave-$n$-users-out cross-validation technique. At each fold, $n$ users were designated for the test set, while the remaining users were split between the training ($90\%$) and validation ($10\%$) sets. Specifically, for the DOMINO dataset we set $n = 1$, while for the Extrasensory dataset, consistently with other works in the literature~\cite{cruciani2020feature}, we set $n = 5$. 
We simulated several data scarcity scenarios by downsampling the available training data at each fold. At each iteration, we used the test set to evaluate the recognition rate in terms of macro F1 score. To ensure robust results, we conducted each experiment $5$ times, calculating the average F1 score.

\subsection{Results}

Figures~\ref{fig:domino-main} and ~\ref{fig:extrasensory-main} show, for both datasets, the impact of infusing \acronym\ knowledge (using the $k$ value leading to the best results) compared to the baselines. 
As expected, infusing knowledge from \acronym\ significantly outperforms the purely data-driven \textit{No knowledge} baseline in data scarcity scenarios. At the same time, \acronym\ reaches competitive results to the \textit{Ontology} baseline, with the great advantage of significantly reduced human effort. 
We also observe that, when the training set includes a sufficient number of labeled samples, the contribution of knowledge infusion is significantly reduced. In these cases, it is likely that most domain constraints are learned from the training data itself. 

Consistently with existing works~\cite{cruciani2020feature,arrotta2024semantic}, the in-the-wild nature of ExtraSensory leads to relatively low recognition rates. Moreover, on this dataset, increasing the percentage of training data slightly degrades the recognition rate of both knowledge infusion approaches.
The reason may be due to the underlying NeSy model, in which raw sensor data may overshadow the infused knowledge when the training set is large. Since DOMINO is significantly smaller, we do not observe the same phenomenon on this dataset.

For both datasets, knowledge infusion is particularly effective for those activities that significantly depend on the context (e.g., using the elevator, moving by car, moving on public transportation), while its benefits are reduced for those activities where context influence is limited (e.g., sitting, walking).

The best results on DOMINO were achieved with low values of $k$ (i.e., ranging from $0$ to $0.25$) and thus a high number of examples. This is due to the fact that, in this dataset, there is a high number of possible context conditions. Thus, the LLM benefits from more examples describing the required task.
On the ExtraSensory dataset, the best results are associated with higher values of $k$ (i.e., ranging from $0.25$ to $0.95$) and thus selecting a smaller number of examples. In this case, since the number of possible context conditions is significantly lower compared to DOMINO, a few examples (similar to the input context) provide the LLM model a better guidance on the required task.

%


%% file: sections/discussion.tex

Our results show that by using pre-trained LLMs instead of ontologies in NeSy HAR systems, we can reach similar (and sometimes better) recognition results.
Then we may wonder why LLMs are a more promising approach for future real deployment of these systems, since we already have some ontologies like the ones used in our work, while for LLMs some prompt engineering work is required.

Considering the extension of these systems to a large class of human activities and different context situations, there are clear limitations for ontologies. To the best of our knowledge, we are not aware of publicly available ontologies offering a comprehensive coverage of all possible human activities and context situations. 
Hence, significant human effort would be required to extend and adapt the ontology to new datasets with different sets of activities and context data. 
Indeed, extending an ontology means defining complex knowledge relationships between activities and contexts, it requires skills in the logic-based formalism model (e.g., OWL2 in the case of ontologies) and significant expertise in the HAR domain. This task is also usually assigned to a single knowledge engineer or small team with high risks of incompleteness.
In our case, adapting \acronym\ to a new dataset only requires using natural language to adjust the system message on the target activities, extending the Context2Text module to map new contexts to natural language descriptions, and generating a new pool of examples. 

However, a significant disadvantage of LLMs compared to ontologies is the absence of real semantic reasoning, since the output is based on data-driven text generation. Hence, there may be contradictions and/or model hallucinations that we would not experience by using rigorous logic systems. For instance, in one instance where the user was moving slowly, the model considered as possible \textit{standing} with the following motivation:\textit{``Possible, as the user is moving at a relatively slow pace''}.
While hallucinations may be reduced by more advanced LLMs models (e.g., GPT-4), they may still occur. One of the advantages of NeSy models is that they can robustly cope with noise in the infused knowledge~\cite{arrotta2024semantic}.

knowledge infusion needs to cope with possibly noisy information.




%% file: sections/conclusion.tex
In this paper, we introduced \acronym: a novel method based on Large Language Models (LLMs) to retrieve common-sense knowledge about human activities to be infused in Neuro-Symbolic (NeSy) context-aware HAR models. We showed the effectiveness of \acronym\ in data scarcity scenarios with an extensive experimental evaluation using a state-of-the-art NeSy model. Our results show that LLMs may effectively replace logic-based models in NeSy systems to reduce human effort.

We have several plans for future work. First, in this work we considered a general LLM incorporating knowledge from many domains. We want to investigate how to specialize the LLM in the HAR domain. This may be obtained with RAG or fine-tuning techniques. 
%
Our investigation also shed some light on the context information released to external LLM services that may expose the subjects to privacy threats. We plan to investigate if more lightweight and open-source LLMs running on trusted edge machines may provide the same knowledge infusion quality.
